\documentclass[runningheads]{llncs}

\usepackage{eccv}

\usepackage{eccvabbrv}

\usepackage{graphicx}
\usepackage{booktabs}
\usepackage{mathrsfs}
\usepackage{enumerate}
\usepackage{multirow}
\usepackage{color}
\usepackage{threeparttable}
\usepackage{pifont}

\usepackage[accsupp]{axessibility}  

\usepackage[pagebackref,breaklinks,colorlinks,citecolor=eccvblue]{hyperref}

\usepackage{colortbl}
\usepackage{orcidlink}
\definecolor{mygray}{gray}{.9}

\hypersetup{
  colorlinks   = true, 
  urlcolor     = cyan, 
  linkcolor    = red, 
}

\begin{document}

\title{Boosting Consistency in Story Visualization  with Rich-Contextual Conditional Diffusion Models}
\titlerunning{Rich-Contextual Conditional Diffusion Models}

\author{Fei Shen\and Hu Ye\and Sibo Liu\and Jun Zhang\thanks{Corresponding author}\and Cong Wang \and Xiao Han\and Wei Yang}
\institute{Tencent AI Lab \\
\email{\{ffeishen, huye, siboliu, junejzhang, xvencewang, haroldhan, willyang\}@tencent.com}}


\authorrunning{Fei Shen et al.}


\maketitle

\begin{abstract}
Recent research showcases the considerable potential of conditional diffusion models for generating consistent stories. 
However, current methods, which predominantly generate stories in an autoregressive and excessively caption-dependent manner, often underrate the contextual consistency and relevance of frames during sequential generation.
To address this, we propose a novel \textbf{R}ich-contextual \textbf{C}onditional \textbf{D}iffusion \textbf{M}odel\textbf{s} (RCDMs), a two-stage approach designed to enhance story generation's semantic consistency and temporal consistency.
Specifically, in the first stage, the frame-prior transformer diffusion model is presented to predict the frame semantic embedding of the unknown clip by aligning the semantic correlations between the captions and frames of the known clip.
The second stage establishes a robust model with rich contextual conditions, including reference images of the known clip, the predicted frame semantic embedding of the unknown clip, and text embeddings of all captions. 
By jointly injecting these rich contextual conditions at the image and feature levels, RCDMs can generate semantic and temporal consistency stories.
Moreover, RCDMs can generate consistent stories with a single forward inference compared to autoregressive models.
Our qualitative and quantitative results demonstrate that our proposed RCDMs outperform in challenging scenarios.
The code and model will be available at
\href{https://github.com/muzishen/RCDMs}{https://github.com/muzishen/RCDMs}.

\end{abstract}

\section{Introduction} \label{sec:intro}
Story visualization ~\cite{li2019storygan, makeastory, arldm} aims to depict a continuous narrative through multiple captions and reference clips. 
It has profound applications in game development and comic drawing. 
Due to the technological leaps in generative models such as generative adversarial network (GAN)~\cite{creswell2018generative, qiao2019mirrorgan, zhang2019self, xu2018attngan, ding2020ccgan, hong2022depth} and diffusion model~\cite{ ramesh2022hierarchical, zhang2023adding, saharia2022photorealistic, shen2023advancing, ye2023ip}, text-to-image synthesis methods~\cite{zhang2021cross, ramesh2021zero, zhang2023sine, yang2023reco, li2024snapfusion} can now generate visually faithful images through text descriptions. 
However, given multiple captions to generate a continuous story with style and temporal consistency still poses significant challenges.

Existing methods typically employ autoregressive generation and can be broadly classified into GAN-based~\cite{li2019storygan, maharana2021improving, maharana2021integrating, li2022word} and diffusion model-based~\cite{arldm, storydall-e,  makeastory, storygpt}.
GAN-based methods typically comprise a text encoder, image generator, image separation, and story discriminator. 
These components work together to maintain the consistency of the entire sequence of images.
However, the images generated by these methods often display distorted objects, mismatched semantics, and localized blurring, especially when creating images from complex scene descriptions.
Subsequently, GAN-based methods~\cite{maharana2021improving, maharana2021integrating} progressively focus on generating more consistent images by improving the performance of the text encoder, such as caption enhancement~\cite{maharana2021improving}, structured text parsing~\cite{maharana2021integrating}, and ID attention mechanism~\cite{chen2022character}.
While these methods~\cite{maharana2021improving, liu2021generative, maharana2021integrating} can generate images that satisfy the requirements of character consistency, they often struggle to maintain a consistent style and capture realistic scene details. 
Furthermore, since the adversarial nature of the min-max objective, GAN-based methods can be prone to unstable training dynamics, which limits the diversity of the stories.

\begin{figure}[t]
  \centering
  \includegraphics[width=\linewidth]{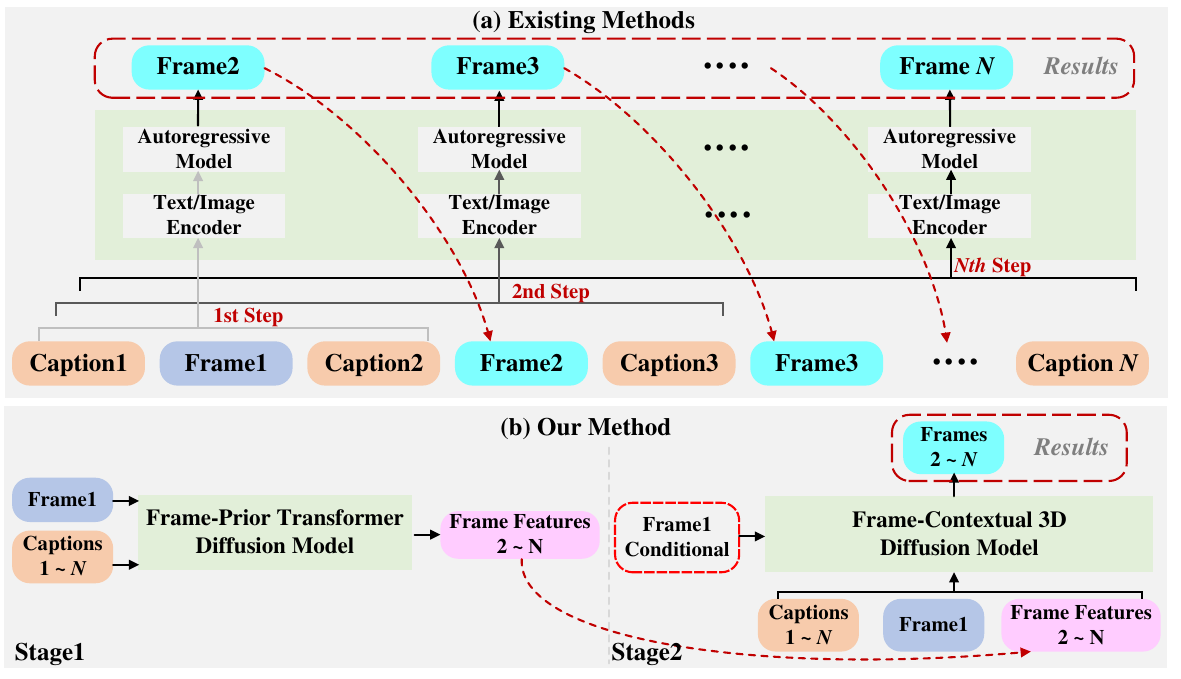}
  \caption{(a) Existing methods, which employ autoregressive models and rely on the current caption for guidance, suffer from weak conditioning, leading to a decrease in the consistency of the generated story.
(b) RCDMs initially predict the frame-contextual information at the feature level, then simultaneously infuse image-level and feature-level contextual information to generate coherent stories in a single forward inference.
  }
  \label{fig:example}
  \vspace{-10pt}
\end{figure}

Advanced diffusion models like Imagen~\cite{imagen} and Stable Diffusion~\cite{sd} have recently demonstrated unprecedented text-to-image synthesis capabilities. 
Story visualization based on diffusion models~\cite{arldm, storydall-e,  makeastory, storygpt} generates better consistent images through a multi-step denoising process, using the features of the current caption and historical frames as conditions. 
However, these methods~\cite{arldm, storydall-e, makeastory, storygpt} only consider clip information at the feature level due to the injection of the current caption and the known clip into the model via the inherent cross-attention module, overlooking the frame information of known at the image level and text information of the unknown clip. 
Besides, from Fig.~\ref{fig:example} (a), GAN-based and diffusion model-based autoregressive generation methods~\cite{arldm, storydall-e, xiao2023survey, lee2022autoregressive} depend on the current caption for frame-by-frame forward generation. 
This reliance can lead these methods to easily overlook the rich contextual information within the captions and the consistency of the story.

This paper presents \textbf{R}ich-contextual \textbf{C}onditional \textbf{D}iffusion \textbf{M}odel\textbf{s} (RCDMs) to tackle the issues above through two stages, as shown in Fig.~\ref{fig:example} (b).
Firstly, we propose a frame-prior transformer diffusion model to predict the semantic embeddings of frames within an unknown clip. 
This prediction task is significantly less complex than directly generating consistent a story. 
It enables the model to focus solely on the task at the semantic feature level, thereby circumventing the rigorous consistency requirements associated with story generation at the image level.
The frame-prior transformer diffusion model takes the frames of the known clip and all captions as conditions, leveraging a combination of multiple cascaded transformer blocks and frame attention blocks to predict the frame semantic embedding of the unknown clip.
In the second stage, instead of over-reliance on caption conditions, we integrate the text embeddings of all captions, the frame semantic features of the unknown clip, and the images of the known clip to serve as a rich contextual guide for generating frames in the unknown clip.
Based on the above rich-contextual conditional, we devise a frame-contextual 3D diffusion model to generate consistent stories by jointly infusing conditions at both the image and feature levels.
Besides, RCDMs can generate consistent stories with a single forward inference compared to autoregressive models.
The main contributions are briefly summarized as follows:
\begin{itemize}
  \item[$\bullet$] We propose a frame-prior transformer diffusion model that, by using the frames from the known clip and all captions as conditions, predicts the semantic embeddings of frames in an unknown clip, thereby providing a rich semantic feature for the next stage.
  \item[$\bullet$] We devise a frame-contextual 3D diffusion model that jointly infuses image-level and feature-level rich-contextual conditional to generate stories with stylistic and temporal consistency.
  \item[$\bullet$] We conduct comprehensive experiments on two datasets to demonstrate the competitive performance of proposed RCDMs. Additionally, we perform a user study to evaluate the superiority of RCDMs qualitatively.
\end{itemize}

\vspace{-15pt}
\section{Related Work} 
\vspace{-5pt}
\subsection{Text-to-Image Synthesis}
Recent advances in text-to-image synthesis mainly focus on generative adversarial networks (GANs)~\cite{qiao2019mirrorgan, xu2018attngan, zhang2017stackgan,tao2022df, zhang2021cross} and diffusion models. 
For example, MirrorGAN~\cite{qiao2019mirrorgan} introduces the concept of redescription, progressively enhancing images' diversity and semantic consistency by focusing on local words and global captions as conditions.
Similarly, AttnGAN~\cite{xu2018attngan} synthesizes fine details by calculating fine-grained image-caption matching losses in different image sub-regions, focusing on relevant words in natural language descriptions.
Moreover, StackGAN~\cite{zhang2017stackgan} uses a two-stage approach, first sketching the shape and color of objects based on the caption and then generating images based on the results of the first stage and the caption. 
On the other hand, XMC-GAN~\cite{zhang2021cross} employs an attention self-modulation generator and a contrastive discriminator to capture and reinforce the correspondence between captions and images, maximizing their mutual information to generate images. 
While these methods~\cite{qiao2019mirrorgan, xu2018attngan, zhang2017stackgan, zhang2021cross} can generate images that meet semantic requirements, GAN training heavily depends on selecting hyperparameters and can easily lose scene details.

Unlike GANs, diffusion models~\cite{song2020score, wang2024v, ho2020denoising, rombach2022high, wang2024ensembling} do not suffer from mode collapse and potentially unstable training and can generate more diverse images. For example,
DALL-E2~\cite{ramesh2022hierarchical} introduces a two-stage model. Initially, it generates learned CLIP image embeddings in an autoregressive fashion, using text descriptions as a guide via a prior model. Subsequently, a diffusion model-based decoder is employed to generate semantically consistent images based on these embeddings.
Following this, Imagen~\cite{saharia2022photorealistic} aims to augment text comprehension capabilities by utilizing a large transformer language model, thereby enhancing image-text alignment and the fidelity of the samples.
Stable Diffusion~\cite{rombach2022high} presents a novel approach of applying diffusion models in the latent space, striking an almost perfect balance between simplification and detail preservation.
However, these text-to-image methodologies~\cite{saharia2022photorealistic, ge2023expressive, rombach2022high, wu2023harnessing} primarily concentrate on aligning individually generated images with their corresponding text descriptions. They overlook the crucial elements of style and temporal consistency across multiple frames, which are essential for effective story visualization.

\subsection{Story Visualization }
StoryGAN~\cite{li2019storygan} is a pioneer in story visualization, proposing a sequential conditional GAN framework with a context encoder that can dynamically track the story flow and a story-level discriminator.
DuCo-StoryGAN~\cite{maharana2021improving} presents a dual learning framework to enhance the semantic consistency between the story and the generated images by improving the captions.
VLC-StoryGAN~\cite{maharana2021integrating} introduces a Transformer-based recursive architecture to encode structured text input through constituency parsing trees while using common sense information to enhance the structured input of the text, making it more in line with the sequence structure of the story. 
Similarly, Word-Level SV~\cite{li2022word} also focuses on text input and proposes a discriminator with fused features and extended spatial attention to improve image quality and story consistency.
Besides, VP-CSV~\cite{chen2022character} devises a two-stage framework that predicts character tokens and the remaining tokens to better ensure character consistency.

Story visualization~\cite{arldm, ahn2023story, storydall-e, storygpt} has achieved great development during these years, especially with the unprecedented success of diffusion model. 
For example, StoryDALL-E~\cite{storydall-e} explores story visualization through an autoregressive Transformer, focusing on full model fine-tuning based on pre-trained models and parameter-efficient adaptive adjustments based on prompts.  
AR-LDM~\cite{arldm} proposes an autoregressive latent diffusion model based on historical captions and generated images to align and enhance story consistency.
Similarly, Story-LDM~\cite{makeastory} proposes an autoregressive diffusion framework with a visual memory module to capture character and background context implicitly.
However, the above autoregressive methods~\cite{li2019storygan, ahn2023story, storydall-e, storygpt}, including GAN-based and diffusion model-based, rely on the current caption for guidance and suffer from weak conditioning, leading to a decrease in the consistency of the generated story.

\section{Method} 

The proposed \textbf{R}ich-contextual \textbf{C}onditional \textbf{D}iffusion \textbf{M}odel\textbf{s} (RCDMs) aims to generate consistent stories by jointly infusing rich contextual conditions at both the image and feature levels. 
In this section, we introduce the following three aspects: preliminaries~(Section~\ref{pre}), frame-prior transformer diffusion model~(Section~\ref{prior}), and frame-contextual 3D diffusion model~(Section~\ref{contextual}).

\subsection{Preliminaries}\label{pre}
Diffusion models represent a category of generative models trained to reverse a diffusion process. This process systematically adds Gaussian noise to the data through a fixed Markov chain of timestep ${t}$, while concurrently training a denoising model to generate samples starting from the Gaussian noise.
To learn such a diffusion model ${\epsilon}_\theta$ parameterized by ${\theta}$, for each timestep ${t}$, the training objective usually adopts a mean square error loss ${L_{DM}}$, as follows,
\vspace{-5pt}
\begin{equation}
L_{DM} = \mathbb{E}_{{x}_{0}, {\epsilon},{c}, t} \| {\epsilon} - {\epsilon}_\theta\big({x}_t, {c}, t\big)\|^2,
\vspace{-5pt}
\end{equation}
where ${x}_{0}$ represents the real data with an additional condition ${c}$.
The timestep ${t}$ of diffusion process is denoted by ${t\in [0, T]}$. The noisy data at ${t}$ step, ${x}_t$, is defined as $\alpha_t{x}_0+\sigma_t{\epsilon}$, where $\alpha_t$ and $\sigma_t$ are predefined functions of $t$ that determine the diffusion process. Once the model ${\epsilon}_{\theta}$ is trained, images can be synthesized from random noise through an iterative process.

During the sampling stage, the predicted noise is calculated based on the predictions of both the conditional model ${\epsilon}_{\theta}({x}_t, {c}, t)$ and the unconditional model ${\epsilon}_{\theta}({x}_t, t)$ via classifier-free guidance~\cite{ho2022classifier}.
\vspace{-5pt}
\begin{equation}
\hat{{\epsilon}}_{\theta}({x}_t, {c}, t) = {w}{\epsilon}_{\theta}({x}_t, {c}, {t})+(1-{w}){\epsilon}_{\theta}({x}_t, {t}).
\vspace{-5pt}
\end{equation}
Here, ${w}$ is the guidance scale used to adjust condition ${c}$. 

\begin{figure}[t]
  \centering
  \includegraphics[width=0.95\linewidth]{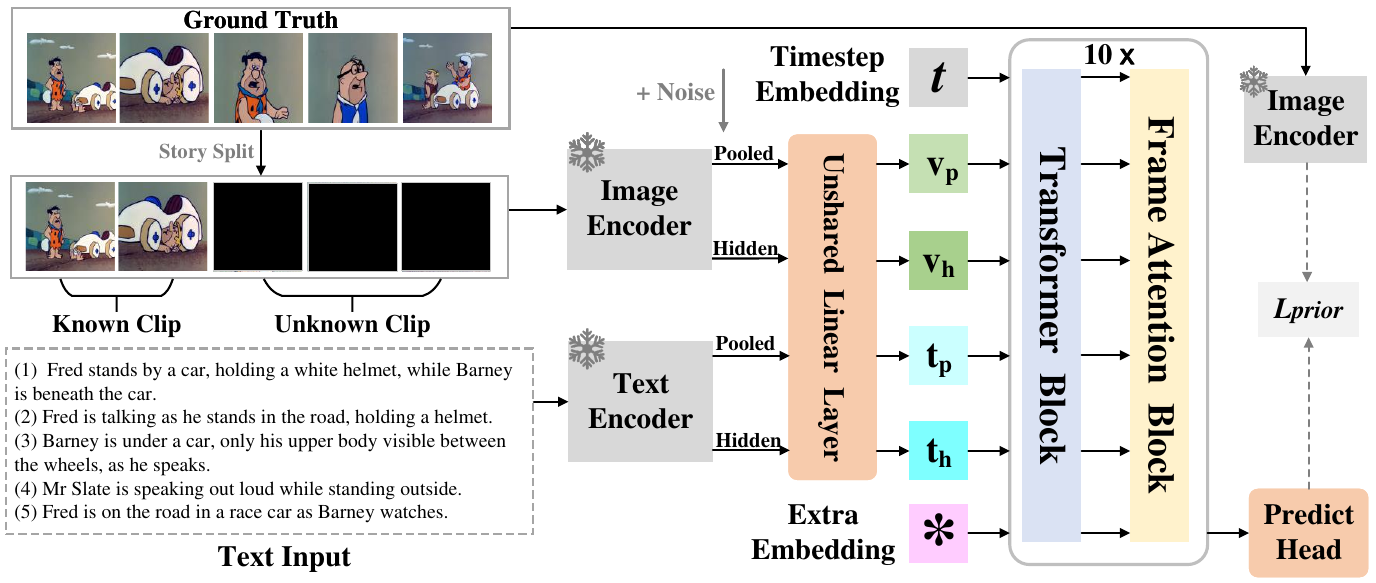}
     \vspace{-.4cm}
  \caption{Illustration of the frame-prior transformer diffusion model. 
  The frame-prior transformer diffusion model predicts  the frame semantic embeddings of unknown clips by aligning the semantic correlations between the captions and frames of known clips.
  }
  \vspace{-15pt}
  \label{fig:prior}
\end{figure}

\subsection{Frame-Prior Transformer Diffusion Model}\label{prior}
In the first stage, we propose a frame-prior transformer diffusion model to predict the frame semantic embeddings of unknown clips by aligning the semantic correlations between the captions and frames of known clips.
As shown in Fig.~\ref{fig:prior}, it is composed of a frozen image encoder, a frozen text encoder, an unshared linear layer, and a stack of multiple transformer blocks and frame attention blocks. 
Here, we utilize the pooled representation extracted from the CLIP image encoder as the frame semantic embeddings for unknown clips.
Our choice is inspired by the capability of CLIP~\cite{radford2021learning}, which is trained on a large-scale dataset of image-text pairs through contrastive learning. This enables it to encapsulate a rich variety of image content and stylistic information, pivotal in steering the subsequent process of story synthesis.

Specifically, we first split the ground truth into known and unknown clips and introduce noise into the unknown clips. 
Subsequently, all clips (known and unknown) and all captions are fed into the frozen image encoder and the frozen text encoder, respectively.
We then use an unshared linear layer to obtain the pooled visual representation $v_p$ and hidden visual representation $v_h$ of the image and pooled textual representation $t_p$ and hidden textual representation $t_h$ of the caption. 
For consecutive frames, we input them as a 4D tensor $x \in \mathbb{R}^{b \times f \times n \times d}$, where $b$, $f$, $n$, and $d$ represent the batch size, temporal length, token length and each token dimension, respectively. 
When the internal feature map passes through the transformer block, the temporal length $f$ is reshaped to the batch size $b$ and ignored, allowing the model to process each frame independently. 
We reshape the feature map back into a 4D tensor after the transformer block, frame attention block reshape $n$ into $b$ to learn and maintain the temporal consistency of the story, and then reshape it back after the module while ignoring the temporal length. 
Inspired by~\cite{blattmann2023align, guo2023animatediff}, the frame attention block consists of several self-attention modules.
This allows the model to guide self-attention along the temporal length $f$, effectively capturing the dynamic content within the narrative.
Besides, we add an extra embedding to represent the unnoised frame semantic embedding of the known clip to be predicted.

Following unCLIP~\cite{ramesh2022hierarchical}, the frame-prior transformer diffusion model is trained to predict the unnoised frame semantic embedding directly rather than the noise added to the frame embedding. 
The training loss $L_{prior}$ of frame-prior transformer diffusion model ${x}_\theta$ is defined as follows,
\begin{equation}
L_{\text{prior}} = \mathbb{E}_{{x}_0, {\epsilon}, {v_p}, {v_h}, {t_p}, {t_h}, {t}} \left\| {x_0} - {x_\theta} \left({x_t}, {v_p}, {v_h}, {t_p}, {t_h}, {t}\right) \right\|^2 .
\end{equation}
Once the model learns the conditional distribution, the inference is performed according to Eq.~\ref{eq:prior_test}, as follows, $w$ is the guidance scale.
\vspace{-5pt}
\begin{equation}\label{eq:prior_test}
\hat{{x}}_\theta\left({x_t}, {v_p}, {v_h}, {t_p}, {t_h}, {t}\right)=w {x}_\theta\left({x_t}, {v_p}, {v_h}, {t_p}, {t_h}, {t}\right)+(1-w) {x}_\theta\left({x}_t, t\right).
\vspace{-5pt}
\end{equation}

\begin{figure}[t]
  \centering
\includegraphics[width=0.95\linewidth]{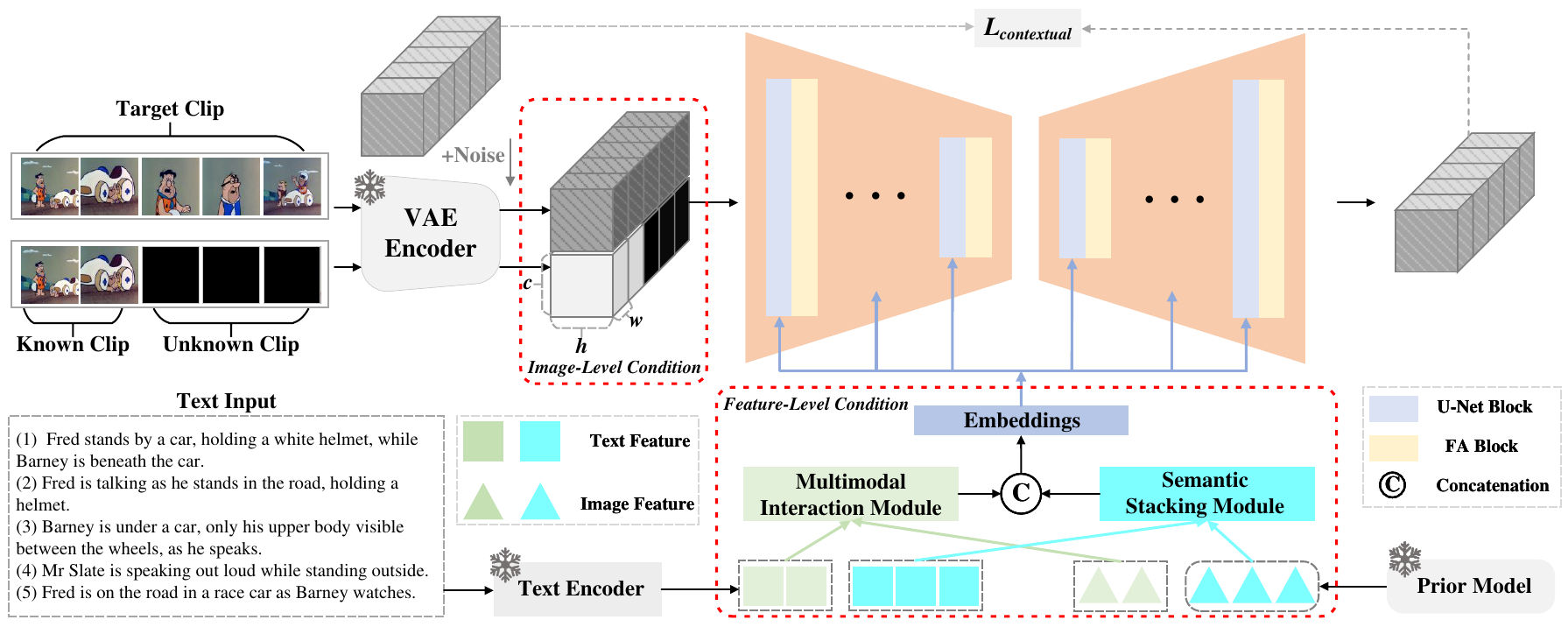}
  \caption{Overview of the frame-contextual 3D diffusion model.
  The frame-contextual 3D diffusion model infuses both image-level and feature-level context information to generate stories with stylistic and temporal consistency.
  }
  \label{fig:framework}
\vspace{-0.2cm}
\end{figure}

\subsection{Frame-Contextual 3D Diffusion Model}\label{contextual}
In the second stage, we propose a frame-contextual 3D diffusion model that utilizes a variety of rich contextual conditions, including reference images from the known clip, the anticipated frame semantic embedding from the unknown clip, and text embeddings from all captions, to generate consistent stories.
Therefore, our method can generate style and temporal consistency stories by jointly injecting these rich contextual conditions at the image and feature levels.
From Fig.~\ref{fig:framework}, the frame-contextual 3D diffusion model comprises a frozen VAE, a model stacked with multiple U-Net blocks and frame attention (FA) blocks, a multimodal interaction module, and a stacked semantic module. 
Here, the design of the FA block is the same as in the previous stage, and focus on the correlation between frames to maintain consistency across frames.

\noindent\textbf{Image-Level Condition.}
Since the VAE~\cite{kingma2019introduction} enables almost lossless reconstruction, introducing known clip conditions at the image level can steer the story continuity, which has been neglected in previous studies.
Specifically, similar to the previous stage, we first divide the ground truth into known and unknown clips and directly mask the unknown clip. 
The ground truth, known clip, and masked unknown clip are all fed into the frozen VAE encoder to extract latent space features. 
We then concatenate the latent space features of the known clip and masked unknown clip along the width dimension, and simultaneously concatenate them with the latent space features of the ground truth along the channel dimension.
Moreover, if the images in the known clip have 0-pixel values similar to the mask, it can easily mislead the model into thinking these areas need to be generated. 
Therefore, we introduce a single-channel marker symbol that matches the width and height of the concatenated features (omitted in the figure). 
We use 0 and 1 to represent masked and unmasked pixels, respectively. 
This approach helps to reduce model confusion and ensures accurate identification of the generation area.

\noindent\textbf{Feature-Level Condition.}
Existing methods depend solely on the current caption's text embeddings, lacking in maintaining the contextual consistency of the overall narrative.
Contrarily, we incorporate the frame semantic embedding of the unknown clip obtained from the previous stage and the text embeddings of all captions as extra feature-level conditions. 
These rich contextual conditions facilitate the generation of consistent narratives. Furthermore, to enhance the features of the text and frames in both known and unknown segments, we separately introduce a multimodal interaction module and a semantic stacking module.
Specifically, we first employ frozen text and image encoders (omitted in the figure) to extract the text embeddings from all captions, and the image embeddings from known clips, respectively. 
We further divide text embeddings $F_{t}\in \mathbb{R}^{f \times n \times d}$ along the temporal dimension into $F_{t}^{k} \in \mathbb{R}^{f^{k} \times n \times d}$ and $F_{t}^{u} \in \mathbb{R}^{f^{u} \times n \times d}$ based on known/unknown clips to align the text and image modalities, where $f = f^{k} + f^{u}$.
For known clip, we first feed the text embeddings $F_{t}^{k}$ and image embeddings $F_{v}^{k} \in \mathbb{R}^{f^{k} \times n \times d}$ of the known clip into a multimodal interaction module. 
The multimodal interaction module comprises two projection layers and a multi-head cross-attention module. 
The two projection layers are respectively used to project the embeddings of images and text onto the same dimension. 
Then, the projected text and image embeddings are then fed into the multi-head cross-attention module to obtain the interaction features $F_{i}^{k} \in \mathbb{R}^{f^{k} \times n \times d}$ of the known clip.

Considering the discrepancy in the number of tokens and dimensions between the frame semantic embedding and the text embedding, we introduce a semantic stacking module specifically for unknown clips. 
This module aims to enhance the feature-level semantic information of the unknown clip.
The semantic stacking module is composed of a projection layer and a multi-head cross-attention module. 
The projection layer's role is to convert the text embedding of the unknown clip into the same feature dimension as the image embedding of the unknown clip, which was obtained from the previous stage.
Assume that $F^{u}_v \in \mathbb{R}^{f^{u} \times 1 \times d}$ and $F^{u}_t \in \mathbb{R}^{f^{u} \times n \times d}$ respectively are image embedding and projected text embedding of the unknown clip.
To align text and image modal, we first obtain the extracted pooled representation $e^g \in \mathbb{R}^{f^{u} \times 1 \times d }$  of unknown clip captions.
Then, the pooled representation $F^{u}_v$ of the unknown clip is fed into the multi-head cross-attention module to obtain the interaction features of the unknown clip. 
These are then stacked with the hidden representation of the unknown clip along the length dimension to obtain the stacked features $F_{s}^{u} \in \mathbb{R}^{f^{u} \times n \times d}$ of the unknown clip.
Finally, the interaction features $F_{i}^{k}$ and the stacked features $F_{s}^{u}$ are concatenated along the temporal dimension and fed into the U-Net block via inherent cross-attention mechanism in diffusion models.

The loss function $L_\mathrm{contextual}$ of frame-contextual 3D diffusion model according to Eq.~\ref{mse_contextual}, as follows.  Here, ${F}_{I}$ denotes the feature of the image-level conditional.
\vspace{-5pt}
\begin{equation}\label{mse_contextual}
L_\mathrm{contextual}=\mathbb{E}_{{x}_{0},{\epsilon}, {F}_{I}, {F}_{i}^k, {F}_{s}^{u}, t} \| {\epsilon}- {\epsilon}_\theta\big({x}_{t},{F}_{I}, {F}_{i}^k, {F}_{s}^{u}, t\big)\|^2.
\vspace{-5pt}
\end{equation}
In the inference stage, we also use classifier-free guidance according to Eq.~\ref{test_inpainting}.
\vspace{-5pt}
\begin{equation}\label{test_inpainting}
\begin{split}
\hat{{\epsilon}}_{\theta}({x}_t, {F}_{I}, {F}_{i}^k, {F}_{s}^{u}, t) = w{\epsilon}_{\theta}({x}_t, {F}_{I}, {F}_{i}^k, {F}_{s}^{u}, t)+(1-w){\epsilon}_{\theta}({x}_t, t).
\end{split}
\vspace{-5pt}
\end{equation}

\section{Experiments} 

\textbf{Datasets.} We conduct experiments on the FlintstonesSV~\cite{maharana2021integrating} and PororoSV~\cite{li2019storygan} datasets. The former contains 20,132 training sequences and 2,309 testing sequences, encompassing 7 main characters. 
PororoSV includes 10,191 training sequences and 2,208 testing sequences, covering 9 main characters. Following~\cite{arldm, makeastory, storygpt}, for story visualization, we designated the first frame as the source frame and generated the remaining four based on this source frame. 

\noindent\textbf{Metrics.} We conduct a comprehensive evaluation of the model, considering both objective and subjective metrics.
Objective indicators include classification accuracy of characters (Char-Acc) and F1-score of characters (Char-F1), both extracted using InceptionV3.
Additionally, we also consider the fr\'echet inception distance (FID) \cite{fid} score. This metric provides a quality assessment by comparing the distribution of feature vectors derived from both real and generated images.
In contrast, subjective assessments prioritize user-oriented metrics \cite{storygpt}, including the percentage of visual quality, text-image relevance, and temporal consistency.

\begin{table}[t]
    \footnotesize
\renewcommand{\arraystretch}{1.0} 
   \vspace{-.3cm}
\caption{Quantitative comparison of the proposed RCDMs with several SOTA models.}
    \label{tab:sota}
    \vspace{-.5cm}
    \begin{center}
        \begin{threeparttable}
            { 
            \begin{tabular}{c|c|c|c|c}
                \hline
                \rowcolor{mygray}
                \cellcolor{mygray}Datasets&Methods & {FID ($\downarrow$)} & {Char-Acc ($\uparrow$)} & {Char-F1 ($\uparrow$)} \\
                \hline
                \multirow{6}{*}{\begin{tabular}[c]{@{}c@{}}FlintstonesSV~\cite{maharana2021integrating}\end{tabular}}
                & LDM~\cite{sd}  & 82.53 & 9.17 & 22.68  \\
                & StoryGAN~\cite{li2019storygan}    & 74.63   &16.57  & 39.68   \\
                & Story-DALL-E~\cite{storydall-e}    &26.49     & 55.19 &73.43   \\
                & Story-LDM~\cite{makeastory}      & 24.24   & 57.19 & 76.59 \\               
                & AR-LDM~\cite{arldm}     & {19.28}  & 62.58 & 79.25   \\               
                & \textbf{RCDMs (Ours)}     & \textbf{14.96} & \textbf{78.44} & \textbf{85.51}  \\
                \hline
                \multirow{8}{*}{\begin{tabular}[c]{@{}c@{}}PororoSV~\cite{li2019storygan}\end{tabular}}
                 & LDM~\cite{sd}   & 64.52 & 4.31 & 12.74    \\
                & StoryGAN~\cite{li2019storygan}    &49.27     & 9.34 &18.59   \\
                & CP-CSV~\cite{chen2022character}    &40.56     & 10.03 & 21.78   \\
                & DuCo-StoryGAN~\cite{maharana2021improving}    &37.15 &13.97  & 38.01 \\
                & Story-DALL-E~\cite{storydall-e}    &35.90     &27.14 &42.45   \\
                 & Story-LDM~\cite{makeastory}      & 26.64   & 29.19 & 47.56  \\

                &  AR-LDM~\cite{arldm}   &{17.40}  & 35.18 &  55.29 \\ 
                & \textbf{RCDMs (Ours)}      & \textbf{16.25} & \textbf{41.48} & \textbf{59.03}  \\
                \hline
            \end{tabular}
            }
        \end{threeparttable}
    \end{center}
    \vspace{-0.7cm}
\end{table}

\noindent\textbf{Implementations.}
We perform our experiments on 8 NVIDIA V100 GPUs.
Our configurations can be summarized as follows,
(1) In the frame-prior transformer diffusion model, there are 10 layers of cascaded transformer and frame attention blocks, and the width of each transformer block is 2048.
For the frame-contextual 3D diffusion model, we use the pretrained Stable Diffusion V1.5~\footnote{https://huggingface.co/runwayml/stable-diffusion-v1-5} and modify the first convolution layer to adapt additional conditions.
(2)  We employ the AdamW optimizer with a fixed learning rate of $1e^{-5}$ in all stages. 
(3)  Following ~\cite{arldm, makeastory}, we train our models using images of sizes 512 $\times$ 512 for FlintstonesSV and PororoSV dataset.
(4) We employ a data augmentation strategy of dropping images in all two stages, with the drop count ranging from 0 to 5. We substitute the dropped images with black images.
(5) In the inference stage, we use the DDIM~\cite{ho2020denoising} sampler with 20 steps and set the guidance scale $w$ to 2.0 for RCDMs on all stages.
{\color{black} Please refer to Appendix B for more detail.}

\begin{figure}[t]
  \centering
\includegraphics[width=0.95\linewidth]{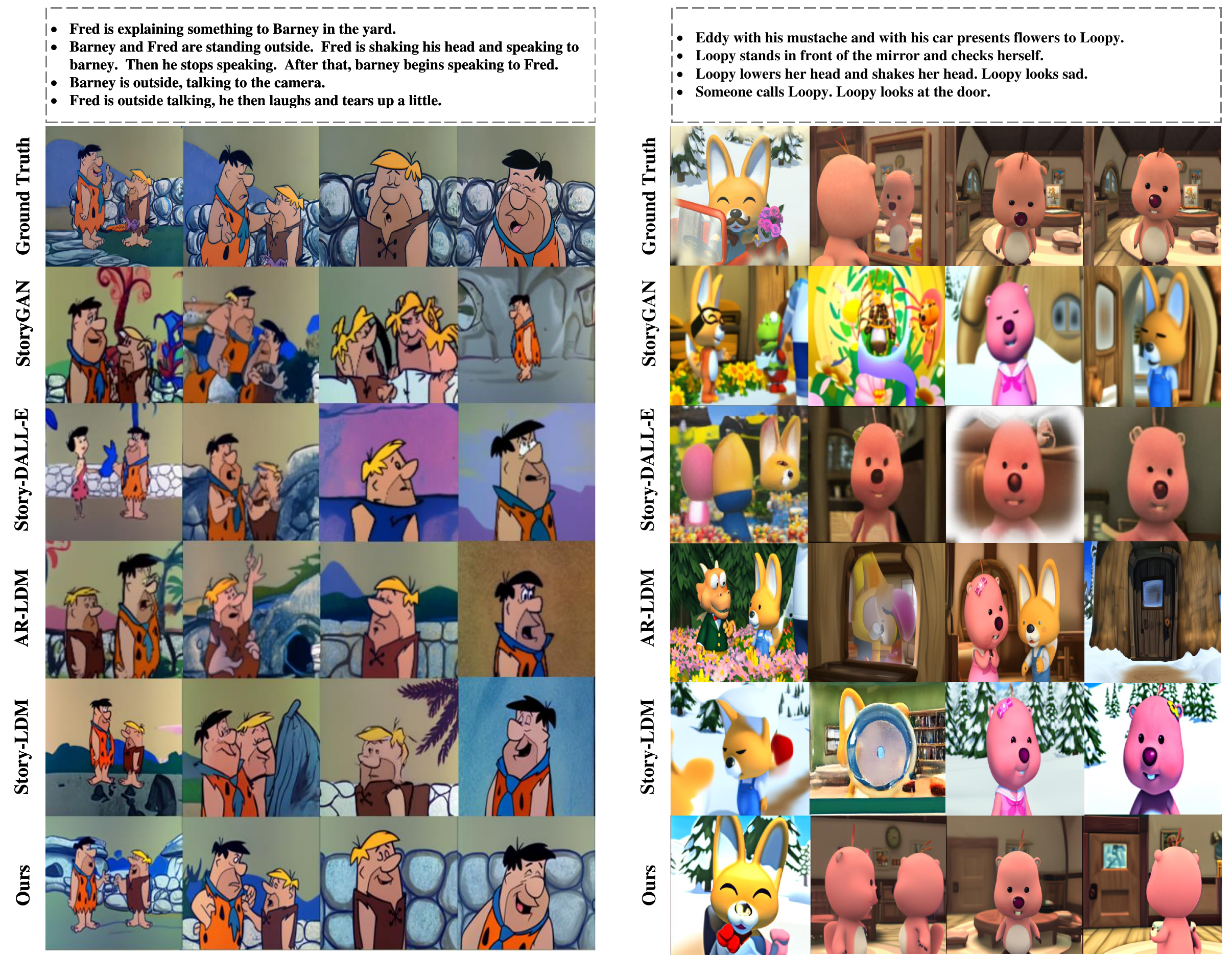}
 \vspace{-0.3cm}
  \caption{Qualitative comparisons with several state-of-the-art models on the FlintstonesSV and PororoSV datasets. {\color{black} {Please see Appendix C for more examples.}}
  }
  \label{fig:sota}
  \vspace{-0.5cm}
\end{figure}

\subsection{Quantitative and Qualitative Results}
We quantitatively compare our proposed RCDMs with several state-of-the-art methods, including  LDM~\cite{sd}, StoryGAN~\cite{li2019storygan}, Story-LDM~\cite{makeastory}, Story-DALL-E~\cite{storydall-e}, AR-LDM~\cite{arldm}, CP-CSV~\cite{chen2022character}, and DuCo-StoryGAN~\cite{maharana2021improving}.

\noindent\textbf{Quantitative Results.}
As shown in Table~\ref{tab:sota}, firstly, since LDM~\cite{sd} generates each image based solely on individual captions, it performs significantly worse than all other methods on three metrics. 
Secondly, compared to GAN and diffusion model methods, our approach outperforms other models on all three metrics in FlintstonesSV. 
For example, compared to StoryGAN~\cite{li2019storygan}, which employs story dynamic tracking, RCDMs score 61.87\% and 45.83\% higher on Char-Acc and Char-F1 metrics, respectively. 
This demonstrates the superiority of proposed RCDMs in understanding story details through all captions and then generating all story images at once, as opposed to StoryGAN, which uses an autoregressive approach to understand captions frame by frame.
Lastly, compared to AR-LDM~\cite{arldm}, which also relies on a diffusion model, RCDMs perform better on the FID metric. 
Even though AR-LDM already scores well, RCDMs show better performance, indicating that injecting more semantic information through the first-stage model can enrich the generation of image details. 
Moreover, on Char-Acc and Char-F1 metrics, RCDMs significantly outperform AR-LDM. 
This is because we not only introduce more semantic information at the feature level but also inject more contextual information at the image level.

The comparison results for PororoSV are summarized in Table~\ref{tab:sota}. 
Notably, consistent with the trend presented in the FlintstonesSV dataset, proposed RCDMs outperform all SOTA methods, achieving the best FID, Char-Acc, and Char-F1.
Specifically, compared to the best-performing GAN method, i.e., DuCo-StoryGAN~\cite{maharana2021improving}, which enhances the current caption to improve text semantic understanding, RCDMs inject context by understanding the complete story semantics.
Similarly, compared to AR-LADM~\cite{arldm}, we surpass it by 6.30\% and 3.74\% on Char-Acc and Char-F1, respectively. 
These results indicate that the simultaneous use of a frame-prior transformer diffusion model to obtain frame semantic information of the unknown clip and the injection of image-level conditions are crucial for understanding and generating stories.

\begin{figure}[t]
  \centering
\includegraphics[width=0.95\linewidth]{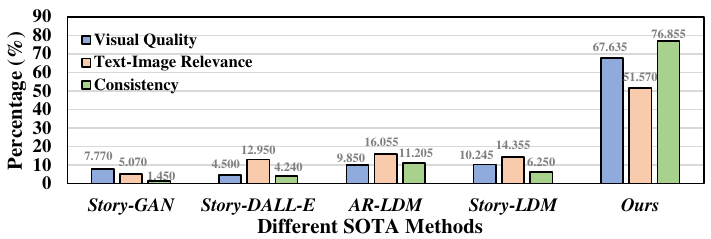}
 \vspace{-.3cm}
  \caption{Results of user study.
      Higher values indicate better performance.
  }
  \label{fig:user_study}
   \vspace{-.7cm}
\end{figure}

\noindent\textbf{Qualitative Results.}
As some methods have yet to be open-sourced, we qualitatively compared RCDMs with StoryGAN~\cite{li2019storygan}, Story-DALL-E~\cite{storydall-e}, AR-LDM \cite{arldm}, and Story-LDM~\cite{makeastory} on the FlintstonesSV and PororoSV datasets. 
As shown in Fig.~\ref{fig:sota}, several conclusions can be drawn from the results:
(1) RCDMs significantly outperform other SOTA methods regarding image quality. 
For example, on the FlintstonesSV dataset,  Story-DALL-E, AR-LDM, and the second frame of Story-LDM, and the third frame of StoryGAN all exhibit scenes and character limbs that do not match.
(2) Regardless of whether it is the FlintstonesSV dataset or the PororoSV dataset, our proposed RCDMs perform best regarding character consistency. 
For example, StoryGAN performs poorly in terms of character consistency on text-image pairs, such as the second and third frames on FlintstonesSV dataset, and the first and fourth frames on PororoSV dataset. A similar situation also occurs on Story-DALL-E.
(3) Only our method can generate story images that reasonably align with the text on complex micro-actions and expressions. 
For example, the text prompt for the fourth frame on the FlintstonesSV dataset is 'laughs and tears up a little,' and the third frame on the PororoSV dataset is 'look sad'.
This can be attributed to our method's ability to enhance the semantic consistency of image-text pairs.
(4) Regarding visual consistency, although Story-LDM integrates an attention-memory module to handle context, it cannot produce a consistent style scene in a complete story. 
In contrast, the results of RCDMs have pleasing visual effects and the ability to maintain temporal consistency, primarily due to RCDMs injecting rich contextual conditions, including reference images of the known clip, the predicted frame semantic embedding of the unknown clip, and text embeddings of all captions at both the feature and image levels.
In summary, our method can always produce more realistic and consistent story images, proving that RCDMs bring significant advantages by introducing rich context conditions.


{\color{black}
\noindent\textbf{User Study.}
The above quantitative and qualitative comparison results demonstrate the substantial advantages of our proposed RCDMs in generating results. 
However, the task of synthesizing story visualization is often human perception-oriented. 
Therefore, we also conduct a user study involving 20 volunteers with computer vision backgrounds. 
The volunteers are asked to choose which method is better regarding visual quality, text-image relevance, and style/temporal consistency in the generated story.
{\color{black}Please refer to Appendix C for more detail.}

As shown in Fig.~\ref{fig:user_study}, the higher the score in the three indicators of this study, the better the performance. RCDMs offer commendable performance on all three fundamental indicators.
For example, the text-image relevance and consistency scores of RCDMs are 51.570\% and 76.855\%, respectively, which are nearly 35.515\% and 65.650\% higher than the second-best model. This demonstrates the significant advantages of RCDMs in multimodal semantic understanding and consistency due to our conditional injection at both the image and feature levels. In addition, our visual quality score is 67.635\%, indicating that participants prefer our method, demonstrating better story visualization quality. More detail refer to appendix~\ref{detail}.
}

\vspace{-0.3cm}

\begin{table*}[t]
\centering
\caption{Ablation study on FlintstonesSV dataset. Here, IC stands for image-level conditional. MIM and SSM, respectively, denote the multimodal interaction module and semantic stacking module.
}
 \vspace{-.3cm}
\begin{tabular}{l||c|ccc|ccc}
\hline

   \multirow{3}*{Settings}  & \multicolumn{4}{c|}{Components} & \multicolumn{3}{c}{FlintstonesSV} \\
  \cline{2-8}
 &\multicolumn{1}{c|}{\multirow{2}*{Stage1}}  &\multicolumn{3}{c|}{Stage2}
  & \multirow{2}*{FID ($\downarrow$)}  & \multirow{2}*{Char-Acc ($\uparrow$)} & \multicolumn{1}{c}{\multirow{2}*{Char-F1 ($\uparrow$)}} \\
  \cline{3-5}
 & & IC & MIM & SSM  &\multicolumn{3}{c}{}\\
  \hline
    LDM~\cite{sd}  &-    &-   &-   &- & 82.53 & 9.17 & 22.68\\
  B0  &\ding{51}    &\color{gray}\ding{55}   &\color{gray}\ding{55}   &\color{gray}\ding{55} & 21.81 & 56.44 & 70.32\\
  B1  &\ding{51}   &\ding{51}   &\color{gray}\ding{55}   &\color{gray}\ding{55} & 19.46 & 61.88 & 78.03\\
  B2  &\ding{51}     &\ding{51}   &\ding{51}   &\color{gray}\ding{55} & 18.36 & 72.53 & 82.06\\
  B3  &\ding{51}   &\ding{51}   &\color{gray}\ding{55}   &\ding{51} &17.94 & 73.58 & 82.87\\
 B4  &\color{gray}\ding{55}   &\ding{51}   &\ding{51}   &\ding{51} &16.51 & 75.73 & 83.96\\
    \textbf{Ours}  &\ding{51}    &\ding{51}   &\ding{51}   &\ding{51} & \textbf{14.96} & \textbf{78.44} & \textbf{85.51}\\
    \hline
\end{tabular}
\label{tab:ablation}
\vspace{-.3cm}
\end{table*}

\subsection{Ablation Study}
We further devise several variants to demonstrate the efficacy of each module proposed in this study. 
All these variants belong to the RCDMs framework but encompass different configurations. 
\textbf{B0} denotes the sole use of the frame-prior transformer diffusion model, devoid of image-level conditions, with feature-level conditions directly infused into the inherent cross-attention of SD using caption features.
\textbf{B1} built on the foundation of B0, incorporates image-level conditions. 
\textbf{B2}, an extension of B1, additionally employs a multimodal interaction module. 
\textbf{B3} also building upon B1, further utilizes a semantic stacking module.
\textbf{B4} indicates that there is no stage1 prior model and in the SSM without stage1, the QKV of attention all originate from the features of the known clips. 

As shown in Table~\ref{tab:ablation}, firstly, compared to LDM~\cite{sd}, B0, which adopts the frame attention module, can significantly improve the consistency of characters and backgrounds. 
Subsequently, when the image-level condition setting is added, B1 is 5.44\% and 7.71\% higher than B0 on Char-Acc and Char-F1, respectively. 
This demonstrates that the injection of known clip images can enhance the model's contextual information.
Secondly, when the MIM and SSM are introduced based on B1, B2 and B3 are 10.65\% and 11.70\% higher than B1 on Char-Acc, respectively. 
At this point, B2 and B3 have achieved highly competitive performance on the FlintstonesSV dataset. 
These results indicate that they also contribute constructively to the success of our RCDMs.
Our proposed RCDMs outperform the B4 setting, highlighting the importance of acquiring prior features in the first stage of our method.
Finally, when the frame semantic embeddings of the unknown clip predicted by the frame-prior transformer diffusion model are added, FID, Char-Acc, and Char-F1 all improve better, especially FID. 
This shows that the frame-prior model can better help RCDMs generate stories with semantic and temporal consistency.

\begin{figure}[t]
    \centering
    \begin{minipage}[t]{0.46\linewidth}
    \includegraphics[width=1\columnwidth]{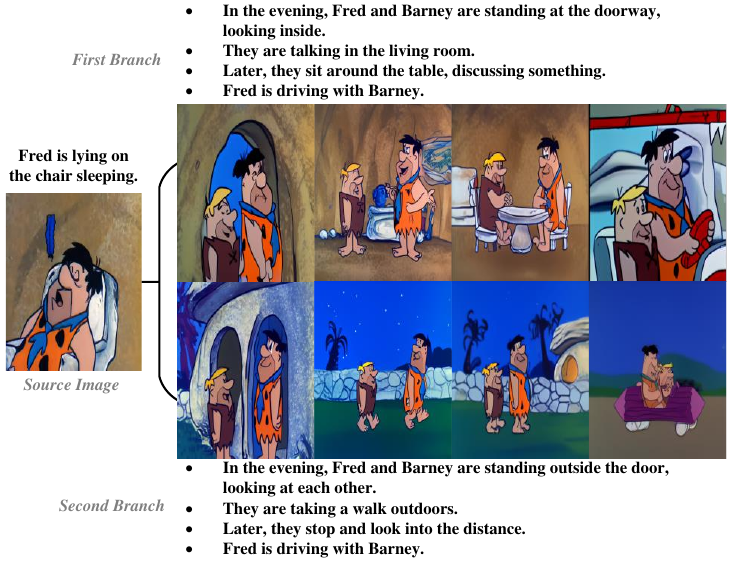}
       \vspace{-.5cm}
    \caption{Branching storyline. Generating consistent stories by different captions.}
    \label{fig:branch}
    \end{minipage}
       \quad
    \begin{minipage}[t]{0.46\linewidth}
    \centering
    \includegraphics[width=1\columnwidth]{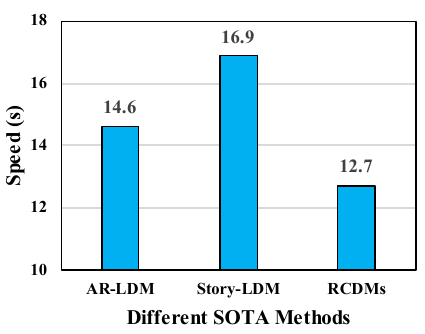}
       \vspace{-.5cm}
    \caption{Inference speed comparison with different SOTA methods.}
    \label{fig:speed}
    \end{minipage}
    \vspace{-0.5cm}
\end{figure}

\vspace{-0.2cm}
\subsection{Additional Results}

\noindent\textbf{Branching Storyline.}
We add an experiment to validate the diversity in generating branching narratives. Specifically, we used one reference image and two sets of distinct captions as the story descriptions. We also incorporated pronouns like ``they" to check if the story could maintain coherence. From Fig.~\ref{fig:branch}, we generated two sets of different story images, and the results demonstrate consistency in style and sequence. Moreover, we observed that for the reference 'they', RCDMs can generate these two characters based on the previously parsed storylines.

\noindent\textbf{Inference Speed.}
We also conduct a comparison experiment on inference speed, including AR-LDM~\cite{arldm} and Story-LDM~\cite{makeastory}, which are also based on the diffusion model.
All experiments are conducted on the same V100 GPU to ensure a fair comparison. 
From Fig.~\ref{fig:speed}, the mean inference time for a story by AR-LDM and Story-LDM is 14.6 seconds and 16.9 seconds, respectively, while proposed RCDMs only take 12.7 seconds, even though it's a two-stage model. 
Since AR-LDM uses a more heavy multimodal model, and Story-LDM employs an attention module with memory storage, both of which slow down the inference speed. 
Notably, while AR-LDM and Story-LDM generate frames one by one using an autoregressive architecture, RCDMs infers all story images in a single forward pass. 
These results demonstrate the architectural superiority of RCDMs.

\begin{figure}[t]
  \centering
\includegraphics[width=0.95\linewidth]{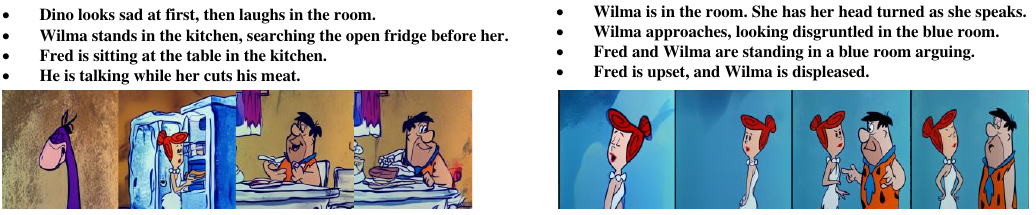}
 \vspace{-.3cm}
  \caption{Qualitative results of RCDMs for caption-only story generation.
  }
  \label{fig:without}
   \vspace{-.5cm}
\end{figure}

\noindent\textbf{Caption-Only Generation.}
RCDMs also support caption-only generation, as it adopts a strategy of randomly dropping images during the training process. 
In contrast, other SOTA methods typically require an additional model to be trained to accommodate this scenario.
Fig.~\ref{fig:without} displays the results generated by RCDMs using only captions as guidance. Due to the lack of open-source weights from SOTA methods for comparison, we can only assess RCDMs' performance independently. The results suggest that RCDMs can generate story images that maintain consistency in style and sequence under different scenarios/characters.

\section{Conclusion}

This paper presents \textbf{R}ich-contextual \textbf{C}onditional \textbf{D}iffusion \textbf{M}odel\textbf{s} (RCDMs) to enhance style and temporal consistency at both the image and feature levels for story visualization.
In the first stage, the frame-prior transformer diffusion model is presented to predict the frame semantic embedding of the unknown clip by aligning the semantic correlations between the captions and frames of the known clip.
The second stage establishes a robust model with rich contextual conditions, including reference images of the known clip, the predicted frame semantic embedding of the unknown clip, and text embeddings of all captions. 
By jointly injecting these rich contextual conditions, RCDMs can generate style and temporal consistency stories.
Both qualitative and quantitative results demonstrate that RCDMs perform well in challenging scenarios.

\noindent\textbf{Limitations.}
Current methods, including RCDMs, typically achieve consistent story generation on a closed-set dataset, which limits the variety of characters and scenes. For future work, we wil explore methods with open-set generation capabilities to allow for a broader range of characters and scenes.

\section{Ethics and Broader Impacts}
This paper presents a synthesis method that creates novel comic narratives from captions and reference images. While there's potential for misuse in fabricating misleading content, this risk is common to all story generation techniques. 
However, the cartoon nature of our outputs significantly reduces ethical concerns related to image and video generation, as they can't be mistaken for reality. 
Moreover, strides have been made in research to detect and prevent malicious tampering. 
Our work aids this domain, balancing the technology's value and the risks of unrestricted access, ensuring the safe and beneficial use of RCDMs.

%
%
\bibliographystyle{splncs04}
\bibliography{main}

\newpage

\appendix

\section*{Supplementary Material}

This supplementary material offers a more detailed exploration of the experiments and methodologies proposed in the main paper.
Section~\ref{nota} provides a series of symbols and definitions for enhanced comprehension. 
Section~\ref{detail} delves deeper into the implementation specifics of our experiments. 
Section~\ref{add_results} presents additional experimental outcomes, including a broader range of supplemented results from the RCDMs method, qualitative comparison examples with state-of-the-art methods, and a detailed explanation of our user studies.

\section{Some Notations and Definitions} \label{nota}

\vspace{-10pt}
\begin{table}[h]
    \renewcommand{\arraystretch}{1.0}
    \centering
    \vspace{-10pt}
    \caption{Some notations and definitions.}
    \begin{tabular}{c|l}
        \hline
        Notation & Definition \\ \hline
        ${x_0}$ & Real image\\
        ${c}$ & Additional condition\\
        ${t}$ & Timestep\\
        ${\theta}$ & Diffusion model\\
        ${\epsilon}$  & Gaussian noise\\
       ${w}$ & Guidance scale\\
        ${v_p}$ &  Pooled visual representation of the frame\\
        ${v_h}$ &  Hidden visual representation of the frame\\
        ${t_p}$ &  Pooled textual representation of the caption\\
        ${t_h}$ &  Hidden textual representation of the caption\\
        ${F_I}$ &  Feature of the image-level conditional\\
        ${F_{i}^{k}}$ &  Interaction feature of the known clip\\
        ${F_{s}^{u}}$ &  Stacking feature of the unknown clip\\
        \hline
    \end{tabular}
    \label{notations}
\end{table}


\section{Implement Details} \label{detail}

Our experiments are conducted on 8 NVIDIA V100 GPUs. We follow the standard training strategies and hyperparameters of diffusion models. 
We utilize the AdamW optimizer with a consistent learning rate of $1e^{-5}$ across all stages. 
The probability of random dropout for condition $c$ is set at 10\%. 
We employ OpenCLIP ViT-bigG/14~\footnote{https://huggingface.co/laion/CLIP-ViT-bigG-14-laion2B-39B-b160k} as the image and text encoder in all stages.
For the frame-prior transformer diffusion model, it consists of 10 transformer blocks, each with a width of 2,048. 
The model is trained for 100k iterations with a batch size of 8, using a cosine noising schedule with 1000 timesteps.
We modify the first convolution layer for the frame-contextual 3D diffusion model to accommodate additional conditions. 
The model is trained for 500k iterations, each with a batch size of 8, and a linear noise schedule with 1000 timesteps is applied.
In the inference stage, we use the DDIM sampler with 20 steps and set the guidance scale $w$ to 2.0 for PCDMs in all stages.

\begin{figure*} [h]
     \centering \includegraphics[width=0.82\linewidth]{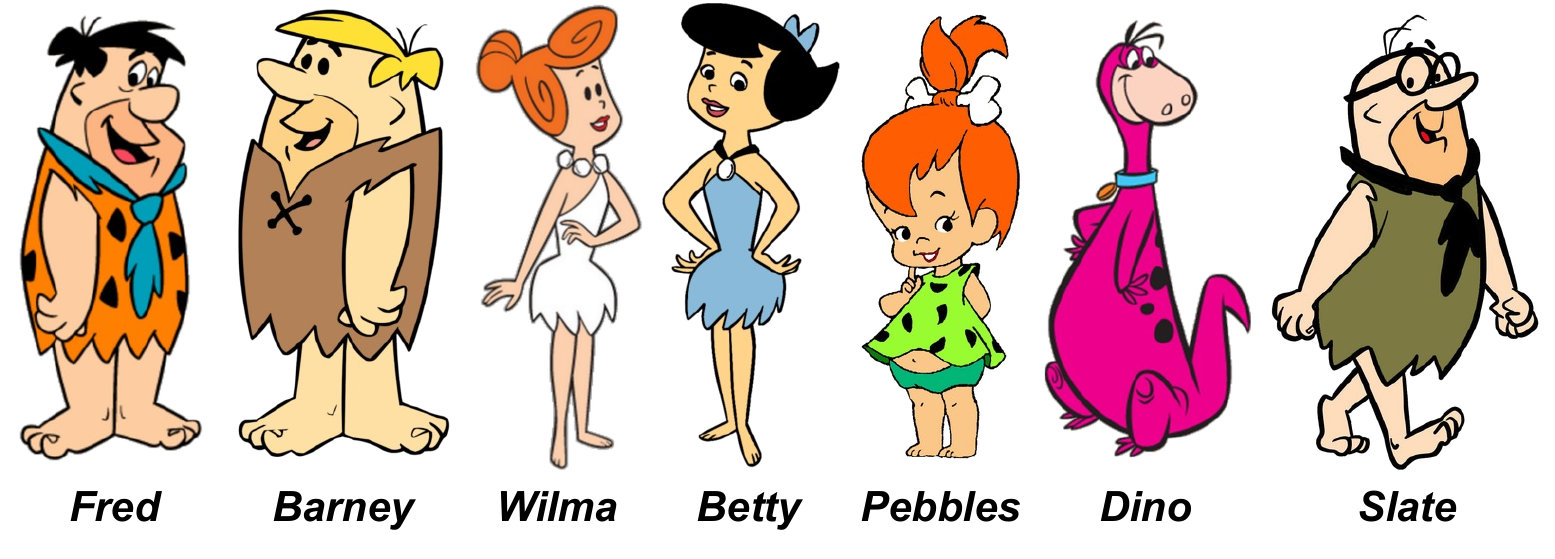}
    \caption{Main character names and corresponding images in FlintstonesSV. The images are from \url{https://flintstones.fandom.com/}.}
    \label{fig:flin_char}
\end{figure*}

\begin{figure*} [h]
\vspace{-20pt}
     \centering \includegraphics[width=0.82\linewidth]{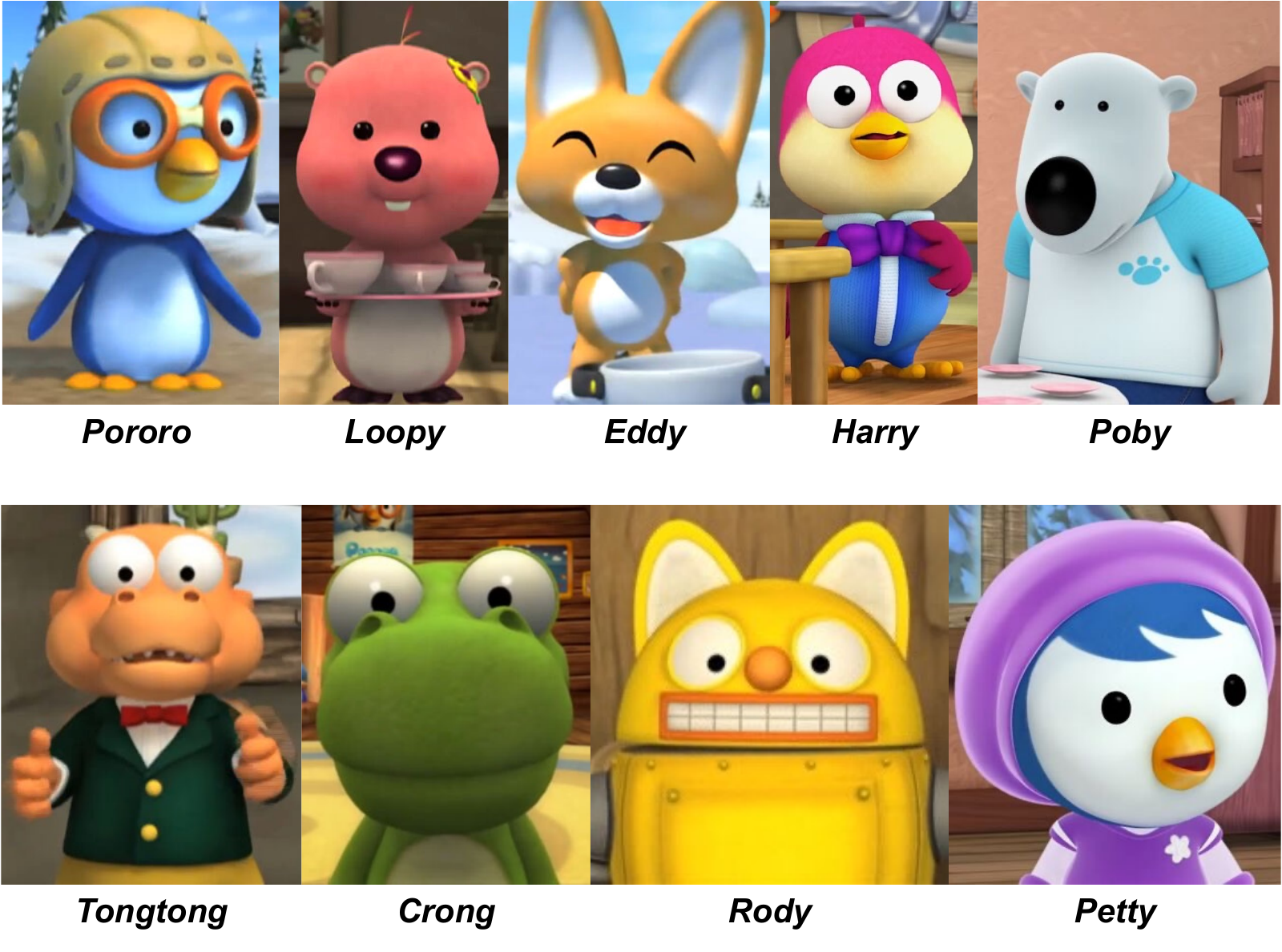}
    \caption{Main character names and corresponding images in PororoSV. The images are from \url{https://pororo.fandom.com/}.
    }
    \label{fig:por_char}
\end{figure*}

\textbf{Dataset.} There are 7 main characters in PororoSV: Fred, Barney, Wilma, Betty, Pebbles, Dino, and Slate. Profile pictures of
them are given in Fig.~\ref{fig:flin_char}.
There are 9 main characters in PororoSV: Pororo, Loopy, Eddy, Harry,
Poby, Tongtong, Crong, Rody, and Petty. Profile pictures of
them are given in Fig.~\ref{fig:por_char}.

\textbf{User Study.}
Details of the user study and need more volunteers.
We carried out a user study using 200 randomly selected sets of stories derived from the FlintstonesSV and PororoSV datasets. These stories were assessed across various dimensions by 100 volunteers.

\section{Additional Results}\label{add_results}
\begin{figure*} [h]
\vspace{-15pt}
     \centering \includegraphics[width=0.85\linewidth]{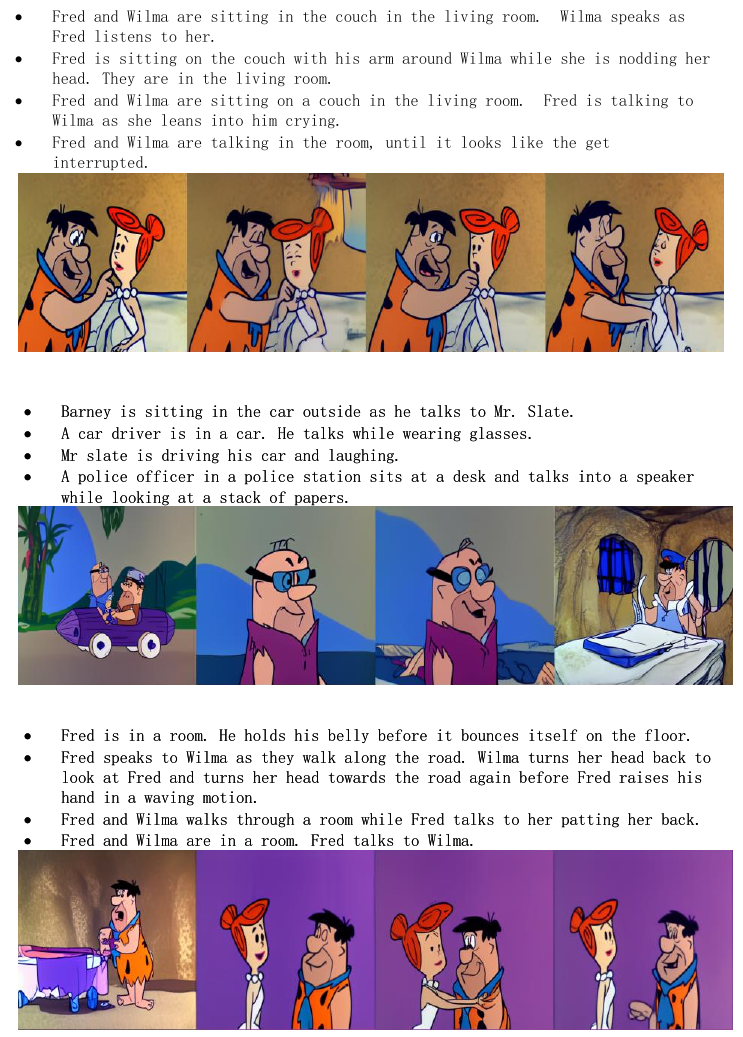}
    \caption{More qualitative results of RCDMs
     on the FlintstonesSV dataset.
    }
\vspace{-23pt}
    \label{fig:ours}
\end{figure*}

\begin{figure*} [t]
     \centering \includegraphics[width=0.9\linewidth]{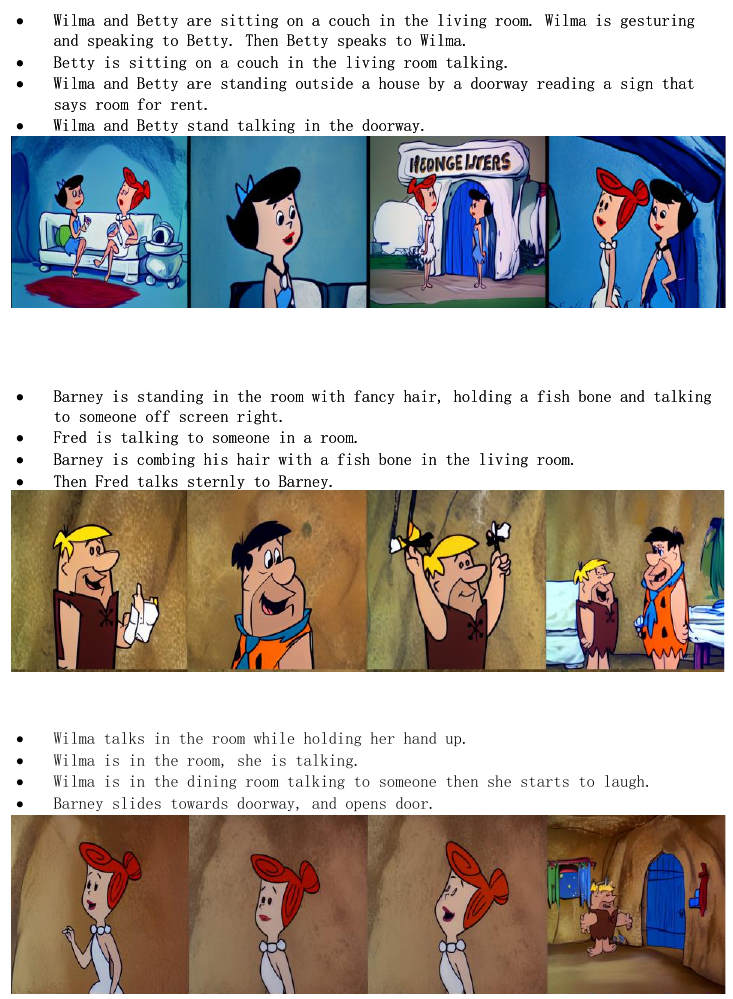}
    \caption{More qualitative results of RCDMs
     on the FlintstonesSV dataset.
    }
    \label{fig:ours1}
\end{figure*}

\begin{figure*} [t]
     \centering \includegraphics[width=0.9\linewidth]{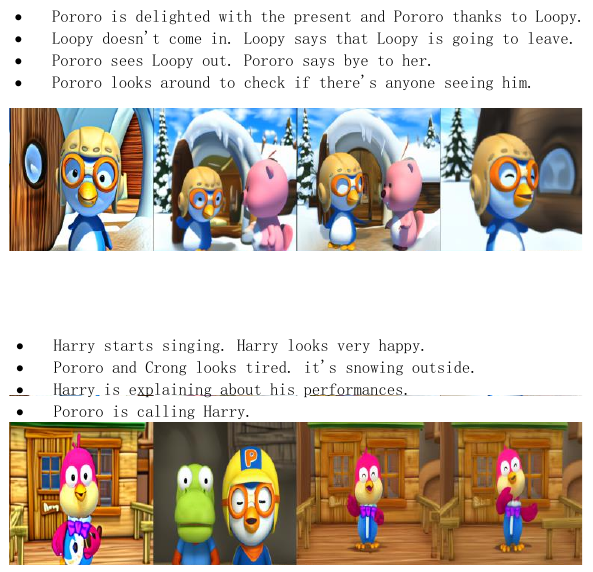}
    \caption{More qualitative results of RCDMs
     on the PororoSV dataset.
    }
    \label{fig:ours2}
\end{figure*}

\begin{figure*} [p]
     \centering \includegraphics[width=0.9\linewidth]{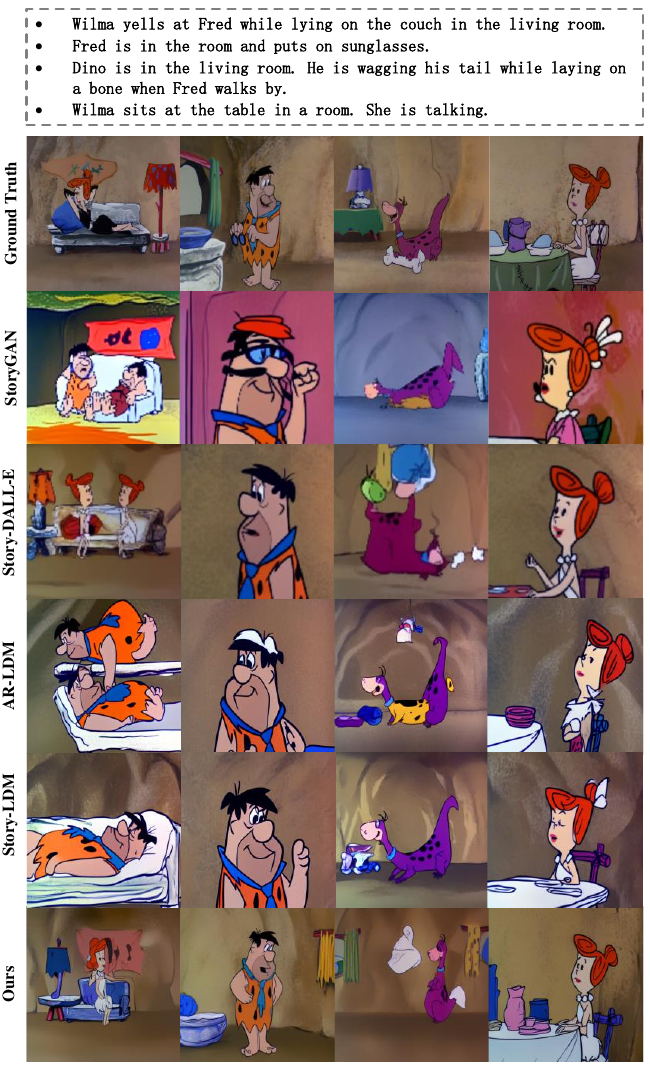}
    \caption{More qualitative comparisons between RCDMs and SOTA methods.
    }
    \label{fig:sota1}
\end{figure*}

\begin{figure*} [p]
     \centering \includegraphics[width=0.9\linewidth]{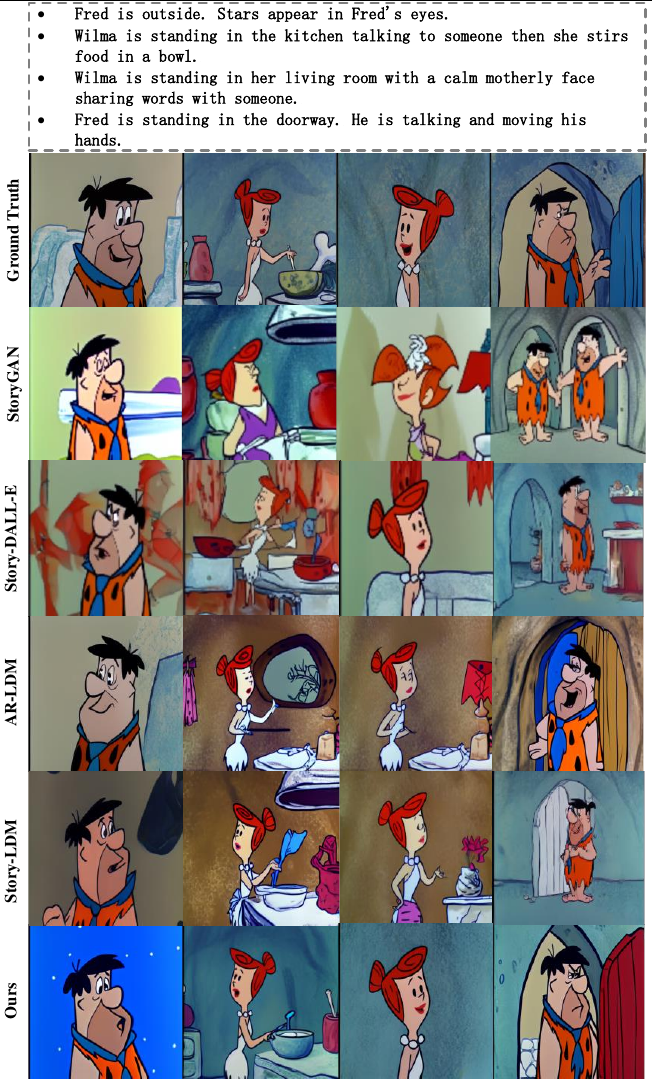}
    \caption{More qualitative comparisons between RCDMs and SOTA methods.
    }
    \label{fig:sota2}
\end{figure*}

\begin{figure*} [t]
     \centering \includegraphics[width=1.0\linewidth]{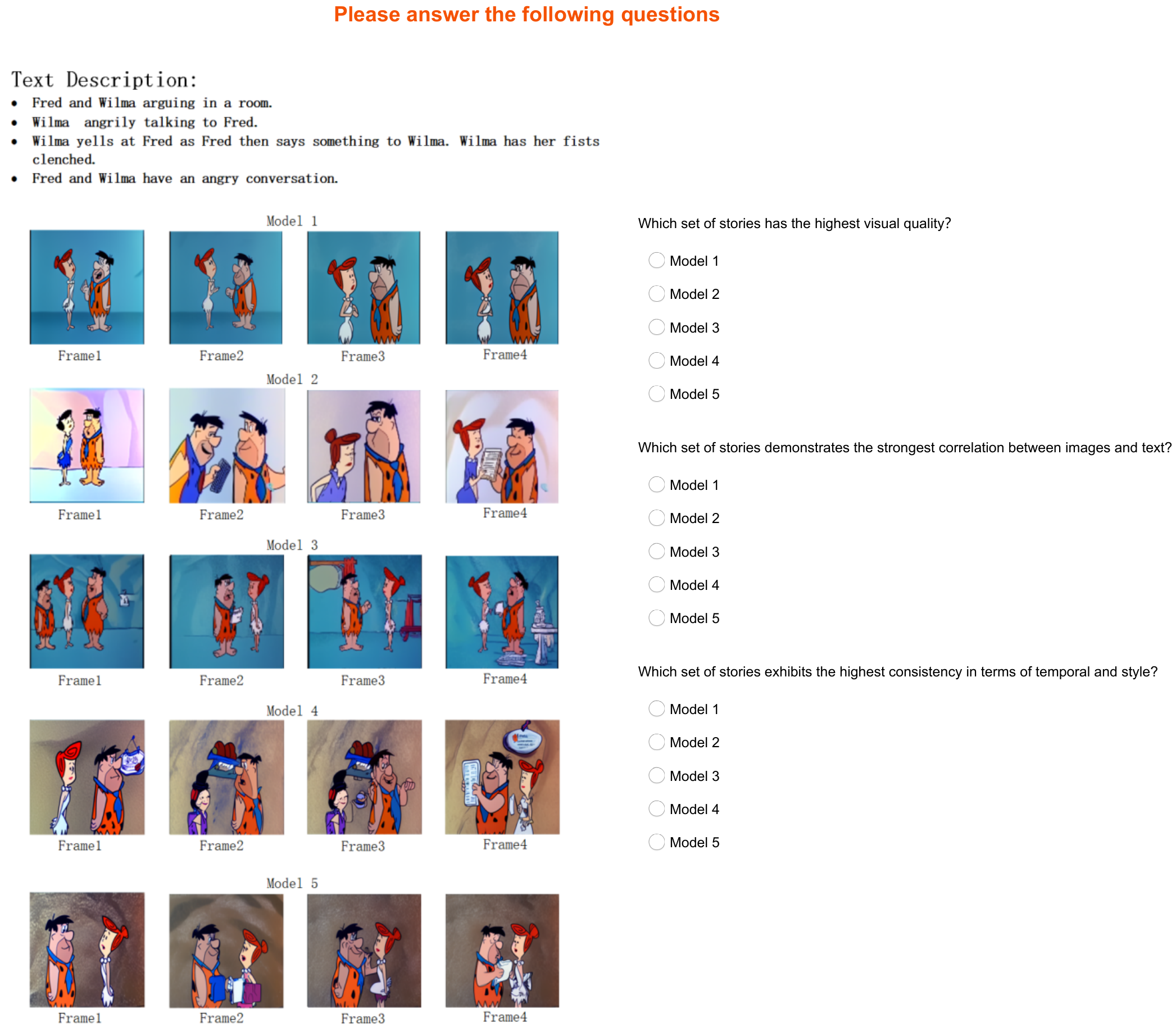}
    \caption{An example question used in our user study for story visualization.
    }
    \label{fig:user}
\end{figure*}

We provide additional examples for RCDMs in Fig.~\ref{fig:ours}, Fig.~\ref{fig:ours1}, and Fig.~\ref{fig:ours2}.
We show additional examples for comparison with the state-of-the-art (SOTA) methods in Fig.~\ref{fig:sota1} and Fig.~\ref{fig:sota2}.
We provide the user study interface as shown in Fig.~\ref{fig:user}.

\end{document}